\documentclass{article}
\usepackage{arxiv}
\usepackage[utf8]{inputenc} 
\usepackage[T1]{fontenc}    
\usepackage{hyperref}       
\usepackage{url}            
\usepackage{booktabs}       
\usepackage{amsfonts}       
\usepackage{nicefrac}       
\usepackage{microtype}      
\usepackage{lipsum}
\usepackage{graphicx}
\usepackage{lscape}
\usepackage{hyperref}
\usepackage{multirow}
\graphicspath{ {./images/} }
\usepackage{algorithm}
\usepackage{algpseudocode}
\usepackage{amsmath}
\usepackage{multirow}
\usepackage{makecell}
\usepackage{tabularx, makecell, booktabs}

\usepackage{float}
\restylefloat{table}
\usepackage{booktabs}

\title{Remote Sensing and Machine Learning for Food Crop Production Data in Africa Post-COVID-19
}

\author{
 Racine Ly \\
  AKADEMIYA2063\\
  Kigali, Rwanda \\
  \texttt{rly@akademiya2063.org} \\
   \And
 Khadim Dia \\
  AKADEMIYA2063 \\
  Kigali, Rwanda \\
  \texttt{kdia@akademiya2063.org} \\
  \And
 Mariam Diallo \\
  AKADEMIYA2063\\
  Kigali, Rwanda \\
  \texttt{mdiallo@akademiya2063.org}\\
}

\makeatletter
\renewcommand\@biblabel[1]{}

\makeatother

\begin{document}
\maketitle


\section{Introduction}
The world is experiencing an unprecedented health crisis throughout the spread of the SARS-CoV-2 (Severe Acute Respiratory Syndrome Coronavirus 2). While the pandemic  (Global Change Data Lab, 2021) on the African continent appears to be less severe than in other geographic regions, the economic impact is significantly more pronounced. The COVID-19 is upending livelihoods, damaging business and government balance sheets, and threatening to reverse sub-Saharan Africa’s development gains and growth prospects for years to come (International Finance Corporation, 2020). The World Bank forecasts that sub-Saharan Africa (SSA) will go into recession in 2020 and that the COVID-19 will cost the region between \$37 billion and \$79 billion in output losses in 2020 alone. The informal sector, a significant source of income and employment, would be the hardest hit. \\
\newline
In the agricultural sector, the COVID-19 threatens to lead to a severe food security crisis in the region, with disruptions in the food supply chain and agricultural production expected to contract between 2.6\% and 7\% (Zeufack et al., 2020). From the food crop production side, the travel bans and border closures, the late reception and the use of agricultural inputs such as imported seeds, fertilizers, and pesticides could lead to poor food crop production performances. Another layer of disruption introduced by the mobility restriction measures is the scarcity of agricultural workers, mainly seasonal workers. The lockdown measures and border closures limit seasonal workers’ availability to get to the farm on time for planting and harvesting activities (Ayanlade \& Radeny, 2020; International Labor Organization, 2020). Moreover, most of the imported agricultural inputs travel by air, which the pandemic has heavily impacted (African Development Bank Group, 2020). Such transportation disruptions can also negatively affect the food crop production system.\\
\newline
It is challenging to establish a comprehensive relationship between the COVID-19 containment measures that countries took and their impacts on food crop production. On the one hand, building such a relationship would imply studies on the impacts of the containment measures on farmers’ and seasonal workers’ mobilities, the prompt reception of seeds, fertilizers, and pesticides for cropping activities. The datasets that would allow the studies mentioned above are not available to our best of knowledge. On the other hand, it is risky to wait until the end of agricultural seasons to have the food crop production statistics and start taking action. Instead, it would be better to have a sense of the most likely food crop production levels before the harvesting period for better planning and early policy actions, and for that goal, data is most needed.\\
\newline
Access to reliable and timely data in the agricultural sector has long been problematic in Africa. Even in regular times, one might experience difficulties in having access to agricultural statistics. The issue is even more pronounced in crisis times, such as the current pandemic, where the data is most needed paradoxically. The most significant effect of the lack of data and analytics is un-informed decision-making processes. In other words, decisions based on anecdotal facts that bring inefficiencies in solving the issues. Much of what we know about agriculture in Africa may no longer be valid, given Africa’s rapid economic transformation, fast urbanization, demographic and climatic changes, and, more importantly, a scarcity of quality data (Christiaensen \& Demery, 2018). In a rapidly changing world, the facts that drive research and policy focus quickly become outdated. The COVID-19 came with the suggestion to improve the African food systems’ resilience. Access to timely, geographically disaggregated, and accurate agricultural statistics can play a significant role in reaching that goal. This is the main scope of this chapter.\\
\newline
This chapter assesses food crop production levels in 2020 – before the harvesting period – in all African regions and four staples such as maize, cassava, rice, and wheat. The production levels are predicted using the combination of biogeophysical remote sensing data retrieved from satellite images and machine learning artificial neural networks (ANNs) technique. The remote sensing products are used as input variables and the ANNs as the predictive modeling framework. The input remote sensing products are the Normalized Difference Vegetation Index (NDVI), the daytime Land Surface Temperature (LST), rainfall data, and agricultural lands’ Evapotranspiration (ET). The output maps and data are made publicly available on a web-based platform, AAgWa (Africa Agriculture Watch), to facilitate access to such information to policymakers, deciders, and other stakeholders.\\
\newline
The chapter is organized as follows: Section 2 provides the underlying conceptual framework that explains the basis for Remote Sensing Products (RSPs) and machine learning use for resilient food systems. Section 3 sets the scene by introducing all variables that have been considered for the predictive model, the methodology used to select the crops for each region, and a short description of the machine learning predictive modeling framework. Section 4 exposes the predicted food crop production for each region and crop. The fifth section proposes recommendations to strengthen African food system resilience through an improved data environment and analytics using emerging technologies.

\section{The basis for the use of RSPs and Machine Learning for Resilient Food Systems}

A resilient food system is determined by its capacity to withstand and recover from disruptions and ensure a sufficient food supply for communities. Another aspect of food systems resilience is the availability of evidence-based technical assistance to help policy and decision-makers more effectively prepare for and respond to shocks. Technology advancements can help with that goal. On the one hand, remotely sensed data via satellite images are now democratized and show high enough spatial resolution to fit a large proportion of agricultural lands across the continent. On the other hand, machine learning techniques offer a way to build robust predictive models relieved from rule-based approaches. This section provides a conceptual framework for understanding the building blocks of our approach to using RSPs and predictive modeling through machine learning techniques for better-informed policies in a time of crisis such as the COVID-19 pandemic. 

\subsection{The basis for the use of RSPs for decision-making in Agriculture}
Near real-time-to-real-time data gathering and analysis is crucial to providing a clear picture of any crisis dynamic and monitoring simultaneous shocks’ effects. The availability of accurate and frequent data that reflects the status on the ground requires significant coordination, collaboration, and robust data systems. In the African food production systems, the level of data quality, frequency, and disaggregation does not allow a thorough analysis of cropping activities, early anomaly detection, and forecasting. RSPs exhibit the promise of significantly reducing the underlying data quality, size, disaggregation, and frequency gaps with earth observation. The RSPs are used in two main ways in agricultural policymaking. First, they provide disaggregated views of agricultural lands and their corresponding biogeophysical parameters. Secondly, they are used to monitor the effects of agricultural policies on the ground. \\
\newline
For the use of RSPs to monitor agricultural lands, earth features’ spectral signature is exploited. For vegetation cover such as agricultural lands, vegetation indices such as the NDVI derived from satellite images assess how healthy crops are by measuring the rate of leaves’ infrared reflection as a proxy for their visible light absorption rate needed for photosynthesis. RSPs provide several other biogeophysical parameters related to food crop production, such as evapotranspiration (Running et al., 2021), land surface temperature (Wan et al., 2015), and the normalized difference water index (NDWI) (Gao, 1996). In general, the combination of specific spectral layers allows determining agricultural land’s biogeophysical status at a community level.\\
\newline
For the use of RSPs to monitor changes on the ground due to agricultural policy, there are some successful examples in other parts of the world. Harnessing MODIS NDVI time-series signals (Lein, 2012) showed how a tax-free agricultural ordinance in 2006 impacted multiple cropping practices adoption in China. (Arvor et al., 2011) derived indices from satellite images to study the relationship between agricultural dynamics in Amazonia during the period 2000-2007 and the region’s existing public policies.\\
\newline
Moreover, satellites revisit the same area several times  a year, allowing practitioners to monitor land use and land cover changes (Li et al., 2020), identify crop taxonomy (Kpienbaareh et al., 2021) and cropping activities (Rezaei et al., 2020), and assess surface water availability (Pekel et al., 2016). Another added value of using RSPs to improve agricultural statistics is its capacity to provide disaggregated information at a pixel level and disentangle the data from conventional administrative entity-based maps. Several weeks (or months) would be necessary to cover the same area with field agents, and still, the results would be less accurate. This capacity to bring the level of detail on the maps down to the community level could allow for targeted response where it is needed the most. However, RSPs alone cannot provide potential future agricultural production and yield. A requirement for that is a predictive modeling framework.

\subsection{The basis for the use of machine learning for decision-making}

General-purpose technologies have been triggering a wide range of innovations globally. The fast pace of technological advances has reduced the cost of technology products and services, encouraged wide adoption, and significantly increased data generated over the last decades. Combined with advances in computer modeling, these advances have opened up a new “technium” of data-driven technologies and machine learning techniques. \\
\newline
Machine learning is a set of techniques particularly suitable for making predictions under certain circumstances. These techniques have capacities to mimic key characteristics that have been attributed to human intelligence, such as vision, speech, and problem-solving. Several papers in the literature have shown how machine learning models outperform human accuracy in some tasks (Buetti-Dinh et al., 2019; Mnih et al., 2013; Silver et al., 2016). This has been possible due to the combination of significant data availability increase, computational power improvements, and algorithmic techniques advancements in the last three decades. The most commonly used supervised learning technique learns how to make predictions as humans by using examples and experience. The approach of transferring human knowledge to machines through sequential steps is being replaced with providing the same data we, as humans, have access to and learn from. Since machines do not have to learn other tasks as do humans, their entire resources are fundamentally oriented towards learning the relationship between the input data and their corresponding outcomes. The result is a faster learning process from the data and better accuracy in a specific task.\\
\newline
As in previous technological revolutions, the most significant impact would be expected in sectors that are not traditional users of these technologies, such as agriculture. Machine learning techniques can support efforts to forecast agricultural productions and yields (Ly \& Dia, 2020; Kaneko et al., 2019), manage natural resources, and reduce uncertainties and risk across the agricultural sector. African farmers are mostly smallholders (Conway et al., 2019) facing significant uncertainties that can lead to poor performance: erratic rainfall, lack of knowledge about biogeophysical parameters and soil water content, and inadequate planting periods. The capacity to forecast agricultural production given these vicissitudes is pivotal for farmers, planners, and policymakers.\\
\newline
Prediction is at the heart of decision-making: however, predictions are just one component in the process. The other decision-making components are judgment, action, outcome, and three types of data: input, training, and feedback (Agrawal et al., 2018). When deciders have access to the same input and training datasets and the same feedback loop, the two key factors that impact their interventions are judgment and the predictions based on the context. While judgment is a subjective concept that depends on background and experience, predictions can follow mathematical formulations; therefore, they can be improved faster and enhance the entire process of designing and implementing informed strategies.

\subsection{Combining RSPs and machine learning for resilient food systems}

The path from RSPs and machine learning to policymaking is not straightforward, especially in Africa. The requirements for sustainable use of RSPs for policymaking in the agricultural sector  necessitate political will, experts in remote sensing and data analysis, and strong institutions with sufficient financial and materials resources to deal with the above tasks. Our rationale for the combination of RSPs and machine learning to build resilient food systems is as follows. RSPs spatial and temporal resolutions allow a disaggregated view of agricultural lands, with several indicators that assess the crop growing conditions at a community level. RSPs as inputs into the machine learning predictive modeling framework is expected to make predictions with reasonable accuracy about food crop production before the harvesting period. Such participate in building a more resilient food system by improving our knowledge of potential agricultural production at the community level. \\
\newline
The availability of ready-to-use biogeophysical RSPs and food crop production forecast maps, in one place and publicly available, would reduce the technical, infrastructural, and institutional barriers that could prevent African countries from exploiting the potential of RSPs and machine learning for resilient food systems, as discussed in sections 2.1 and 2.2. Table 1 shows the underlying techniques and concepts harnessed to provide near real-time biogeophysical data and food crop production maps at the community level for all African countries. The corresponding outputs and outcomes and uplifted constraints for decision-making in food production systems are illustrated.

\begin{table}[H]
\caption {Techniques and concepts and their corresponding outputs and outcomes to harness the combination of RSPs and machine learning for decision and policymaking in the agricultural sector. The focus here is on the food crop production side.} \label{tab:table1} 
    \begin{tabularx}{\linewidth}{l*{2}{X}c}
    \toprule
    \thead{Technique and Concept} & \thead{Outputs} & \thead{Constraints uplifted} \\
    \midrule
    \makecell{The use of RSPs and machine learning to assess \\ policies impacts on food crop production systems} & Use the time-series data provided on the web-based tool to asses if the policy goal are reached or not, and take corrective actions &  \\
    \midrule
    Decision and policy-making based on forecasts\\ and biogeophysical parameters time-series & Use the food crop production disaggregated and forecast map at the community level to plan and strategize based on the scenario provided by the model & All the information based on RSPs and their most likely future outcomes are made available to decision-makers to provide impactful policies. \\
    \midrule
    The use of web-based tool to make datasets and \\maps publicly available & Make the food crop production forecast maps and time-series data used as inputs available in a web-based tool &	Remote sensing products and food crop production forecast at the community level is made publicly available, uplifting the data access constraint.\\
    \midrule
    Forecasts based on the combination of RSPs and \\ML, and third-party data & The combination of inputs variables data and ML method to learn the data patterns and use the data structure learned for future predictions & The technical field level expertise needed for machine learning predictive model is not a constraint anymore \\
    \midrule
    Remote sensing products from satellite images\\ and machine learning techniques & Pre-processed inputs maps such as NDVI, LST, rainfall, ET, production maps as labels, and crop masks and the choice of ML technique &	The lack of technical skills about data processing methods for satellite images is not a constraint anymore for analysts, decision and policymakers. \\
    \bottomrule
    \end{tabularx}
\end{table}

\section{Remotely Sensed Data, Crop Selection, and Predictive Modeling Framework}

Food crop production estimation based on remote sensing can be built through two main approaches: (i) Using remotely sensed data as inputs to Agro-meteorological or plant-physiological models, and (ii) Building a direct mathematical relationship between remotely sensed data and crop production (Huang \& Han, 2014). The first approach is based on “mechanistic” descriptions of crop growth, development, and production simulated through mathematical functions. Such methods have shown satisfactory results but cannot exploit datasets to their full extent due to the constraints coming along the way crop growth phenomena are described with mathematical functions. The second approach usually relies on derived indicators from remotely sensed data and their correlation with crop growth and yield. 

\subsection{Biogeophysical Remotely Sensed Data for Food Crop Production Forecasts}

One of the most known and used parameters to characterize vegetation covers is the NDVI, derived from near-infrared and red bands from multispectral sensors. The NDVI indicator is extensively used to characterized vegetation covers due to its close relationship with several vegetation parameters such as Leaf Area Index (LAI), the fraction of Absorbed Photosynthetically Active Radiation (fAPAR), and green biomass. Many studies have been conducted to predict crop yield from NDVI signals, such as (Liu et al., 2019; Rembold et al., 2013; Rasmussen, 1992,1997). However, there are limitations to using NDVI only as a proxy for crop yield estimation due to its dependencies on the crop, soil, and leaf types. Indeed, even though NDVI is a good proxy for aboveground biomass production, the relationship between biomass and yield varies in time and space (Leroux et al., 2018). Our approach emphasizes the use of several remote sensing products, and such would ensure the use of even more information about crop status than NDVI only.\\
\newline
Several studies conducted in the 1970s have shown that final crop yield can be related to thermal indices (Idso et al., 1977; Smith et al., 1985). Hence, a daytime Land Surface Temperature (LST) layer has been used as a proxy for crop water stress in our methodology. Water availability is also a key component for crop growth and yield; therefore, it is essential when building a crop production model to take it into account. However, for most African countries, agricultural lands are rainfed (Stockholm International Water Institute, 2018). Therefore, rainfall data has been derived from CHIRPS remote sensing products.\\
\newline
Soil water content and its dynamic in the underground agricultural lands is an important parameter to encounter. The underground water is conveyed towards the atmosphere with two main channels: evaporation and transpiration. The former corresponds to the transformation of liquid water bodies into their gaseous state and their release into the atmosphere. For the evaporation process to occur, enough soil moisture, vapor pressure gradient, and a significant amount of energy (600 calories of heat energy for 1g of water) are required. On the other hand, the transpiration mechanism consists of underground water transportation from the soil to plants’ roots, then from the roots to the leaves through the vascular plant tissues, and ultimately, its evaporation into the atmosphere. In the literature (Bhatt \& Hossain, 2019), transpiration is the most desired mechanism since water transportation, through the plant’s internal structure, also carries the soil nutrients from the soil to the plant and prevents the plant’s tissue from overheating. However, measuring the two processes’ contribution in the amount of water in the atmosphere is difficult; therefore, their combined effects are usually measured with the Evapotranspiration (ET) index from remote sensing products. We use the total ET on crop locations as a proxy for measuring the crops’ rate of transpiration, which will inform us about the crops’ health through the effectiveness of the process.\\
\newline
Our production estimation methodology allows us to predict production for one crop in the region of interest before the harvesting period and at the community level. Raster type maps for historical production quantities for 42 crops are publicly available globally from the SPAM  (IFPRI, 2016, 2019, 2020) database. They have been generated with an allocation model with a grid cell size of ten kilometers. Such a map will be used for two purposes: Using the pixel production values as response variables to our model and creating crop masks to target areas where a specific crop is believed to be grown. Table 2 summarized the list of remote sensing products taken as inputs and response variables for our food crop production model and their spatial and temporal characteristics.

\begin{table}[htbp!]
 \caption{Inputs parameters for the food crop production model with their spatial and temporal characteristics.}
  \centering
  \begin{tabular}{lllll}
    \textbf{Inputs parameters} & \textbf{Datasets} & \textbf{Spatial Res} & \textbf{Temporal Res.} & \textbf{Time extent} \\
    \midrule
    NDVI & MOD13A2 & 1 & 16 & 2000-now \\
    LST-Day & MOD11A2 & 1 & 8 & 2000-now \\
    Rainfall & Africa monthly &	5.55 & 30 & 1981-Dec2020 \\
    Evapotranspiration & MOD16A2 & 0.5 & 8 & 2000-now \\
    Production & P & 10 & - & 2000-now \\
  \end{tabular}
  \label{tab:table2}
\end{table}

\subsection{Crop Selection for the Food Crop Production Model}

African farmers are mostly smallholders who grow food for consumption and income. Because of chronic infrastructural and financial issues and difficulties to access agricultural inputs and markets, a relatively low-intensity shock could significantly impact their food security status. Therefore, the knowledge of potential future agricultural production before the harvesting period is essential for planning purposes. The crops should be targeted with their relative importance for a countries’ most vulnerable communities, especially for a major crisis such as the COVID-19 crisis.\\
\newline
The criteria that determine a food crop importance, in this chapter, is a combination of food crop production quantities and food self-sufficiency for a country. Two rankings are performed to identify countries’ most five important food crops. The first one is a ranking of the ten most produced food crops, and the second one is a ranking in terms of food crop self-sufficiency. The analysis relies on agricultural variables that are publicly available on international databases. The production and domestic supply data are available on the United Nations’ Food and Agriculture Organization (FAO) online database for the most recently available years (2014-2018). \\
\newline
For each country, a preliminary list of the ten most produced agricultural commodities in terms of quantities was built. Subsequently, we included the sufficiency aspect through the self-sufficiency ratio defined by the share of food crop consumption to food crop production at the domestic level. The ratio evaluates whether a country produces enough food crops to cover its own needs for each of the ten most produced food crops. Therefore, an agricultural commodity is considered essential for a country if the consumption is greater than the production. The list of the five primary produced and consumed commodities for each country by region are reported in tables A1, A2, A3, A4, and A5 (cf. appendices). 

\begin{table}[H]
\caption {List of selected food crop by African region.} \label{tab:table3} 
    \begin{tabularx}{\linewidth}{l*{2}{X}c}
    \toprule
    \thead{African Regions (No. countries)} & \thead{Food Crops} & \thead{No. of countries where the selected\\ food crop appears in the top 5} \\
    \midrule
    Eastern Africa (14 countries) & Maize & 8 \\
    & Cassava & 8 \\
    & Sugar Cane & 9 \\
    \midrule
    Southern Africa (5 countries) & Maize & 3 \\ 
    \midrule
    Northern Africa (5 countries) & Wheat & 4 \\ 
    \midrule
    Western Africa (16 countries)  & Cassava & 8 \\ 
    & Rice & 9 \\
    & Maize & 7 \\ 
    \end{tabularx}
\end{table}

The most predominant food crops in the top five commodities within countries were selected for the regional level. Table 3 presents the list of selected crops for each region. In eastern Africa, maize, cassava, and sugar cane are selected as the major commodities. Indeed, eight over fourteen countries have, based on our ranking, maize and cassava as the most priority products in terms of production and consumption. Sugar cane is also essential for nine over 14 eastern African countries. In western Africa, three commodities were selected: cassava, rice, and maize. For other regions, cassava, maize, and wheat were identified as essential commodities for central Africa, southern Africa, and northern Africa, respectively.

\subsection{Predictive Modeling Framework}

The supervised learning ANNs method was used to build the food crop production model. The inputs were the first four biogeophysical parameters listed in table 2, and the corresponding outputs are the production values. A pre-processing data stage dealt with building the proper format and splitting the data into training, validation, and testing sets. The learning process is carried out by building the relationship between inputs and response variables with the training dataset. The validation data is used to fine-tune the model. Finally, the testing data allows assessing the model accuracy. An overall arithmetic average of out-of-sample (OOS) root-mean-squared error (RMSE) of 0.9965 was obtained. For all African countries, the smallest OOS RMSE was 0.9864, and the largest was 0.9999553.\\
\newline
Forecasts have been made before the harvesting period (in 2020) for each of the targeted crops (cf. Table 3). The FAO crop calendar was used to identify sowing, growing, and harvesting periods. For the eastern African region, cassava and sugar cane were not considered in this study due to data availability issues. For each country  and crop, the food crop production modeling work starts at the crop greenness onset, which is considered the growing season’s beginning. Most of the biogeophysical parameters were not available at that time; therefore, their historical values were used into a random forest regressor to estimate their future values in the growing season. The latter are then used as inputs to the food crop production model. 

\section{Food Crop Production Forecasts during the pandemic}
The food crop production model was applied to all African countries and the selected crop. Figures 1, 2, and 3 show the 2020 maps predicted production as a share of the 2017 production for rice, maize, and cassava, respectively, for the western African countries. Figures 4, 5, 6, and 7 shows maize production ratio for eastern, southern, and central African regions and wheat for the northern African countries, respectively.\\
\newline
At the regional level, the production quantities in 2020 for each African region are expected to decline compared to 2017 for most of the selected crops. Only the production quantities for cassava in the west and central African regions are expected to increase compared to 2017 by 4.2\% and 28.4\%, respectively. The sharpest decline in production quantities for the three selected crops in the western African region is expected for rice with a decrease close to 12\%, while maize production is expected to decline by close to 5\%. The decline in maize production is expected to be around 1.5\% and 18.6\% for the eastern and southern African regions. Wheat production shows a decline of close to 10\% in the northern African region in 2020 compared to 2017.\\
\newline
According to the United Nations Food and Agriculture Organization (FAO), the six central African countries’ aggregated cassava production was around 47 million metric tons in 2017. The most significant contributor was the Democratic Republic of Congo (66.6\%), followed by Angola (17.9\%) and Cameroon (10.2\%). Our model suggests a total cassava production of close to 60 million metric tons for the same countries in 2020, which corresponds to an increase of 28\% compared to 2017. However, in 2020 the distribution of total production across individual countries is expected to remain the same for Angola while Cameroon shares roughly decreased by half (5.6\%). The Democratic Republic of Congo has a share increase of near +10\%.\\
\newline
In Northern Africa, the ratios between predicted wheat production levels for the 2020 season and actual levels in 2017 show a slight decrease in 2020. On average, the map (figure 5) suggests better wheat production for the 2017 season compared to 2020 for Libya, Tunisia, Morocco, and Algeria. Compared to 2017 production levels, projected wheat production in 2020 in Sudan and Egypt shows a progression of +3.5\% and +2.8\%, respectively, while other countries show a decline. \\
\newline
The aggregated maize predicted production from our model for considered eastern countries is around 28 million metric tons in 2020. Each country’s contribution to the global production was as follows: Ethiopia (27.7\%), Tanzania (20.2\%), Kenya (12.2\%), Zambia (10.9\%), Uganda (10.7\%), Malawi (9.1\%), Mozambique (5\%), Zimbabwe (2.5\%), Rwanda (1.1\%) and Madagascar (0.8\%). In 2017, the production was estimated at around 28.5 million metric tons for the same countries (FAO). There is a slight decrease of 1.6\% in 2020 production estimates compared to 2017. However, some countries such as Ethiopia, Zimbabwe, Uganda, and Zambia show a slight increase in their production of 0.8\%, 13.2\%, 4.6\%, 3.8\%, respectively, compare to 2017.\\
\newline
Figure 6 shows ratios between predicted maize production levels for the 2020 season and actual levels in 2017 in southern Africa. The map suggests disparities in maize production for the 2020 season compared to 2017. Maize production is expected to reduce by 30\% in South Africa, 20\% in Lesotho, 4.9\% in Swaziland, and 0.7\% in Namibia from 2017 to 2020. \\
\newline
Beyond the possibility to monitor crops’ growing conditions through biogeophysical parameters, the combination of remote sensing products and machine learning provides several benefits. As we have seen, the RSPs allow us to bring the disaggregation at the community level while the machine learning techniques help predict food crop production before the harvesting period. The two outcomes mentioned above are valuable assets to strengthen food production systems through improved agricultural statistics and analytics in Africa. However, the path from RSPs and machine learning to policymaking in the agriculture sector requires several building blocks. We discuss three of them in the following section.

\newpage

\begin{figure}[!htbp]
  \caption{The 2020 predicted rice production as a share of the 2017 production. \textbf{Note}: If the ratio is above unity, the 2020 predicted production is expected to be larger than the 2017 production. A ratio smaller than unity means an expected decrease in production in 2020 compared to 2017—data, methodology, and maps’ sources: Authors.}
  \centering
    \includegraphics[width=0.75\textwidth]{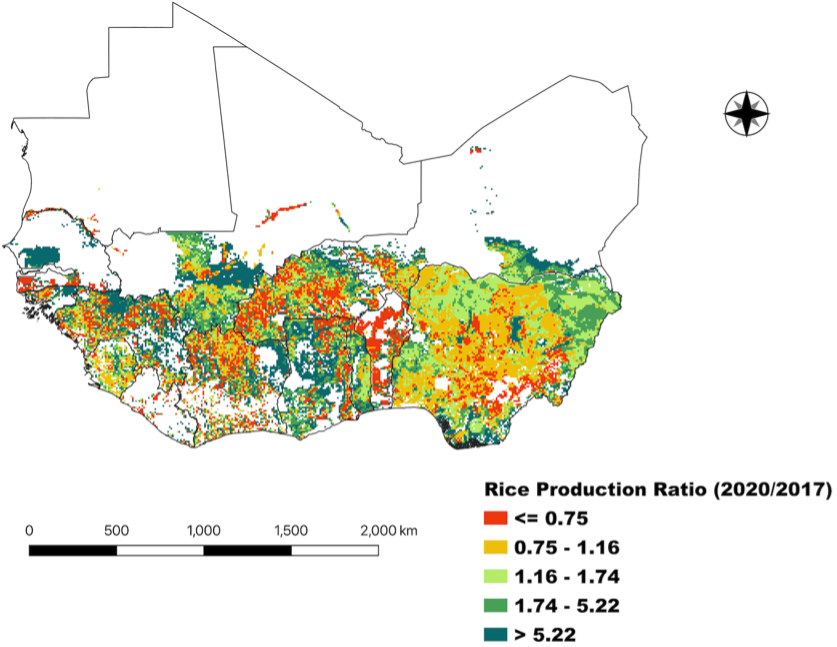}
\end{figure}

\begin{figure}[!htbp]
  \caption{The 2020 predicted maize production as a share of the 2017 production. \textbf{Note}: If the ratio is above unity, the 2020 predicted production is expected to be larger than the 2017 production. A ratio smaller than unity means an expected decrease in production in 2020 compared to 2017—data, methodology, and maps’ sources: Authors.}
  \centering
    \includegraphics[width=0.75\textwidth]{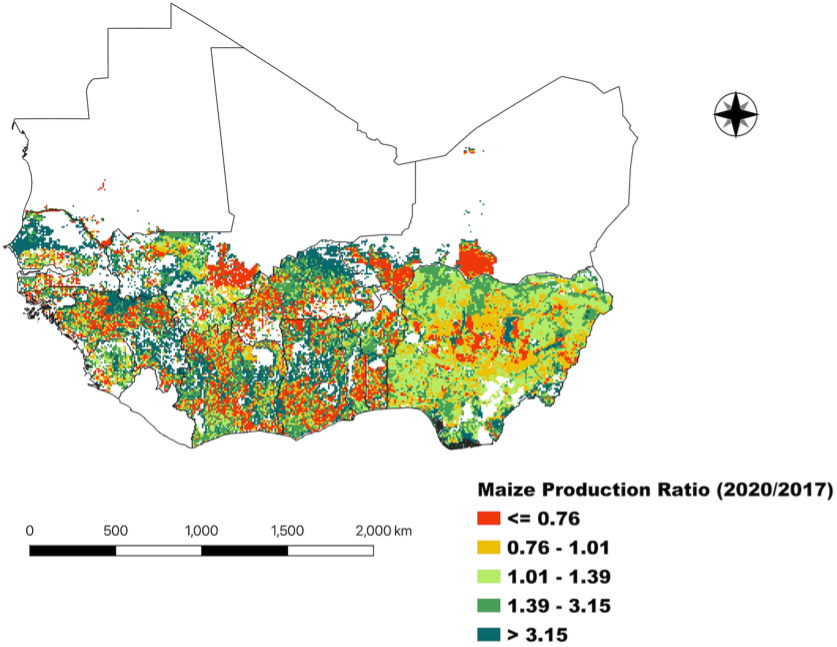}
\end{figure}

\begin{figure}[!htbp]
  \caption{The 2020 predicted cassava production as a share of the 2017 production. \textbf{Note}: If the ratio is above unity, the 2020 predicted production is expected to be larger than the 2017 production. A ratio smaller than unity means an expected decrease in production in 2020 compared to 2017—data, methodology, and maps’ sources: Authors.}
  \centering
    \includegraphics[width=0.75\textwidth]{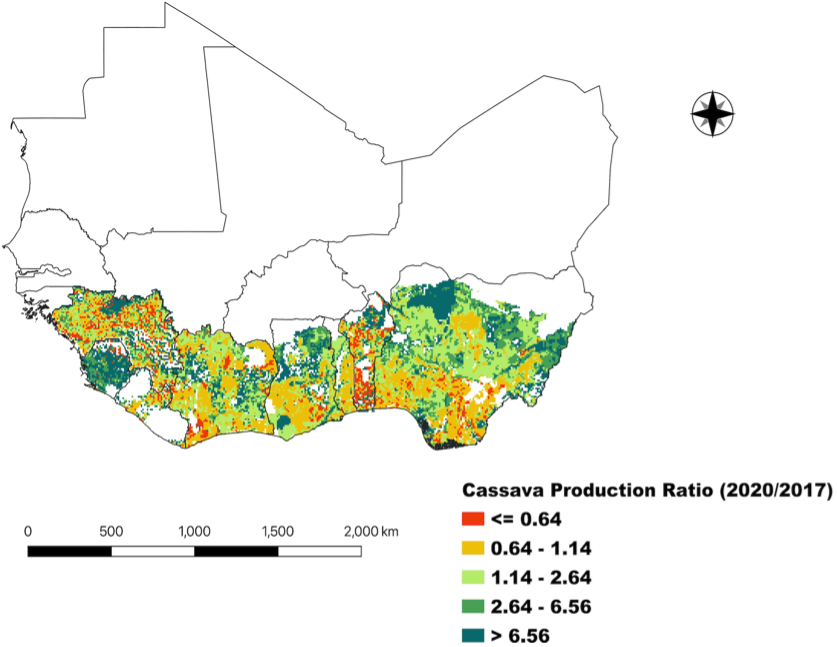}
\end{figure}

\begin{figure}[!htbp]
  \caption{The 2020 predicted rice production as a share of the 2017 production. \textbf{Note}: If the ratio is above unity, the 2020 predicted production is expected to be larger than the 2017 production. A ratio smaller than unity means an expected decrease in production in 2020 compared to 2017—data, methodology, and maps’ sources: Authors.}
  \centering
    \includegraphics[width=0.75\textwidth]{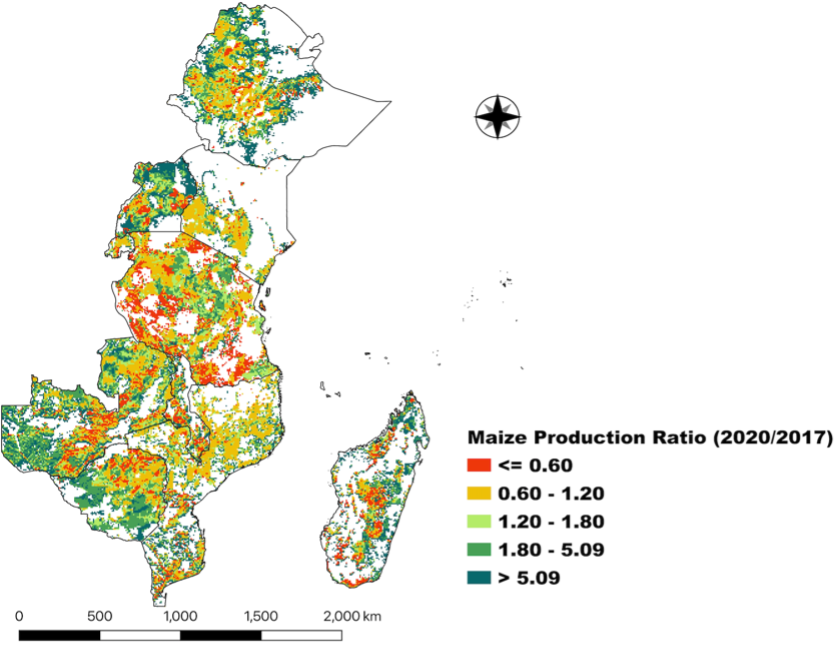}
\end{figure}
    
\begin{figure}[!htbp]
  \caption{The 2020 predicted wheat production as a share of the 2017 production. \textbf{Note}: If the ratio is above unity, the 2020 predicted production is expected to be larger than the 2017 production. A ratio smaller than unity means an expected decrease in production in 2020 compared to 2017—data, methodology, and maps’ sources: Authors.}
  \centering
    \includegraphics[width=0.75\textwidth]{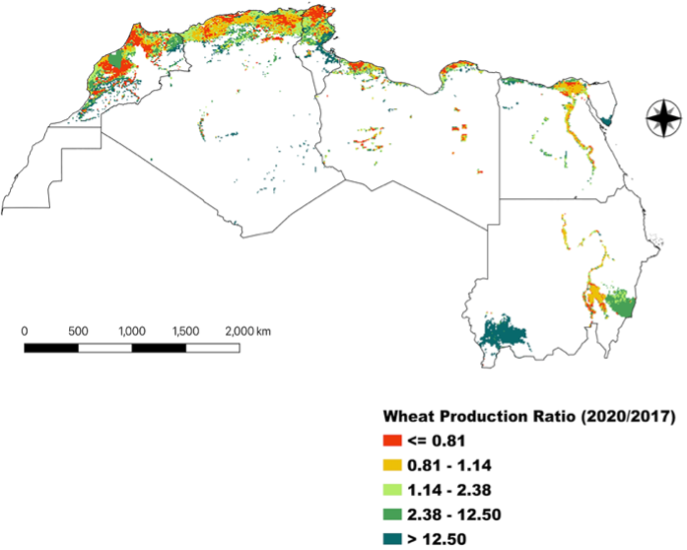}
\end{figure}

\begin{figure}[!htbp]
  \caption{The 2020 predicted maize production as a share of the 2017 production. \textbf{Note}: If the ratio is above unity, the 2020 predicted production is expected to be larger than the 2017 production. A ratio smaller than unity means an expected decrease in production in 2020 compared to 2017—data, methodology, and maps’ sources: Authors.}
  \centering
    \includegraphics[width=0.75\textwidth]{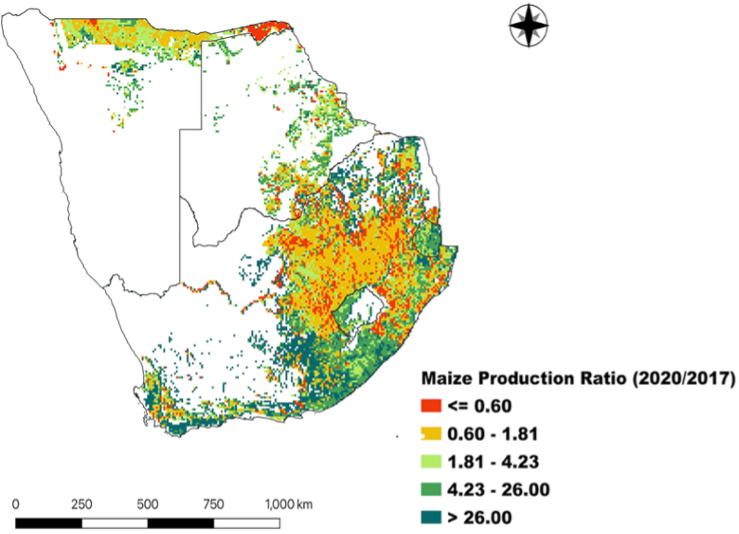}
\end{figure}

\begin{figure}[!htbp]
  \caption{The 2020 predicted maize production as a share of the 2017 production. \textbf{Note}: If the ratio is above unity, the 2020 predicted production is expected to be larger than the 2017 production. A ratio smaller than unity means an expected decrease in production in 2020 compared to 2017—data, methodology, and maps’ sources: Authors.}
  \centering
    \includegraphics[width=0.75\textwidth]{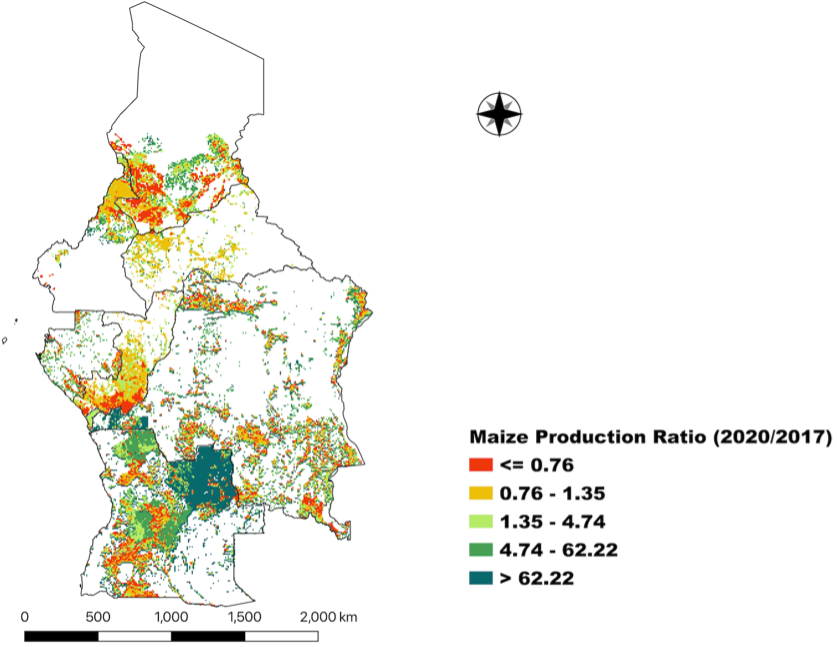}
\end{figure}

\begin{table}[!htbp]
\caption {Total production in 2017 and 2020, and the rate of change for each crop and African region—data source: 2017 production (FAO), 2020 production (authors).} \label{tab:table4}
\begin{tabular}{lllll}
Region                       & Crop    & 2017 production (MT) & 2020 predicted production (MT) & Rate of change (\%) \\
\hline
\multirow{3}{*}{West Africa} & Rice    & 17803495.8           & 15640125.8                     & -12.15              \\
                             & Maize   & 21666866.9           & 20599545.5                     & -4.92               \\
                             & Cassava & 90151658.8           & 93948433.2                     & +4.21               \\
East Africa                  & Maize   & 28539928.7           & 28095011.8                     & -1.55               \\
North Africa                 & Wheat   & 18392407.2           & 16610688.1                     & -9.68               \\
South Africa                 & Maize   & 420814.5             & 342688.341                     & -18.56              \\
Central Africa               & Cassava & 47209110             & 60598537                       & +28.36             
\end{tabular}
\end{table}

\section{Conclusions}

As suggested by the COVID-19 burdens, a robust African agricultural statistics system is much needed for informed and targeted responses and policies. Building a culture of gathering accurate and timely data about features related to food crop production will not only facilitate the production of better policies and monitoring and evaluation mechanisms, but it would also be determinant to increase countries’ level of preparedness for any potential future crisis in the sector: either by identifying it early enough to mitigate its impacts, or by allowing decision and policymakers to manage them better when they occur. This chapter made the case on how emerging technologies such as RSPs and machine learning can be harnessed to provide valuable information for decision-making processes in the agricultural sector. However, the path from a raw satellite image to an informative map is not straightforward; several areas of expertise need to operate at different levels. \\
\newline
Capacity building on emerging technologies such as remote sensing and machine learning should be institutionalized. African governments must create special units where the emerging technologies could be harnessed to inform policies. Moreover, countries would benefit from incentivizing initiatives in the private sector to harness them. However, attracting students into the fields of emerging technologies mentioned above requires public-private solid partnerships and support for entrepreneurship in STEM (Science, Technology, Engineering, and Mathematics) to create jobs and ensure the availability of a critical mass of experts in the corresponding sectors.\\
\newline
In the agricultural sector, across African countries, the data related to food crop production are collected at the season onset and after the harvesting period. The data is mainly collected at the household level with households’ information, the crop type, production quantities, land size, the availability and use of agricultural inputs, post-harvest loss and sailings information. Such a methodology has proven to be sufficient for an extended period. However, given technology improvements and their use for more efficient data gathering processes, there is a need to fully take advantage of the current data-gathering technologies. \\
\newline
Metadata is as essential as the primary data to benefit from recent analytics tools and predictive modeling through machine learning. The metadata helps to contextualize the primary datasets and add more explanatory variables into the predictive model for more robustness. The use of cloud technology, telecommunications, and tablets with embedded optimized forms, could facilitate gathering such third-party information. The cloud would help to store the data and to perform further analysis; the internet connection could help to gather GPS coordinates and inform on the location where the data were gathered (not only at the household level but at the farm itself to allow the analysis of biogeophysical parameters from RSPs). For such to occur, at least three enabling technologies are required: (i) An improved internet connection in rural areas where most of the farms are located; (ii) The inclusion of metadata information gathering into agricultural surveys; (iii) The renewal of data gathering tools to migrate from papers and laptops to tablets that are more suitable for such a task. Such an approach of using emerging and well-established technologies to support a better-quality data gathering about the agricultural sector will progressively require fewer resources due to some data that do not need to be updated from the ground, thanks to the use of remote sensing.\\
\newline
Information asymmetry between researchers and policymakers is longstanding in Africa, especially in the agricultural sector. Moreover, the fast pace of turn-over in offices makes the consolidation of technical knowledge within an institution difficult. For instance, an individual at the national bureau of statistics could be trained to work with remote sensing products and machine learning techniques in a year. The following year, they could be in another ministry, another entity of the same ministry, or transition to another institution. From a general point of view, the training is not lost. However, the corresponding technical capacity moves from one entity to another with the risk of not being used where it is needed the most. \\
\newline
The complex African cropping system makes it difficult to collect accurate and timely data sustainably. On the one hand, such a data scarcity does not allow the type of detailed analysis that decision-making would require in a time of uncertainties. On the other hand, when the data quality and disaggregation requirements are met, the way the knowledge is produced seems not to be digestible to policymakers, especially when emerging technologies are used and are far from reach. One way of closing the gap mentioned above is to appeal to data visualization expertise to transform the data and knowledge from its raw stage to its information stage. Such expertise is not yet well developed across African countries and needs to be built.\\
\newline
The results of this chapter not only support the use of emerging technologies such as RSPs and machine learning techniques to improve agricultural statistics, but they also showed how they could be leveraged to increase African countries’ preparedness to shocks post-COVID-19. The pandemic shows how timely and accurate data are most needed for early action and intervention in the agricultural sector and beyond. Recent technologies must be considered in any part of the data environment – from collection to analysis.\newline

\begin{large}
\textbf{References}
\end{large}

African Development Bank (AfDB). (2020, November). AN EFFECTIVE RESPONSE TO COVID-19 IMPACTS ON AFRICA’S AVIATION SECTOR. \url{https://www.afdb.org/sites/default/files/2020/11 26/afdb_aviation_covid_19_recovery_conference_draft_background_paper_nov2020.pdf.}\\

Agrawal, A., Gans, J., \& Goldfarb, A. (2018). Prediction Machines: The Simple Economics of Artificial Intelligence. Harvard Business Review Press.\\

Arvor, D., Jonathan, M., Meirelles, M. S. P., Dubreuil, V., \& Durieux, L. (2011). Classification of MODIS EVI time series for crop mapping in the state of Mato Grosso, Brazil. International Journal of Remote Sensing, 32(22), 7847–7871. \url{https://doi.org/10.1080/01431161.2010.531783}\\

Ayanlade, A., Radeny, M. COVID-19 and food security in Sub-Saharan Africa: implications of lockdown during agricultural planting seasons. npj Sci Food 4, 13 (2020). \url{https://doi.org/10.1038/s41538-020-00073-0}\\

Bhatt, R., \& Hossain, A. (2019). Concept and Consequence of Evapotranspiration for Sustainable Crop Production in the Era of Climate Change. Advanced Evapotranspiration Methods and Applications, 1–13. \url{https://doi.org/10.5772/intechopen.83707}\\

Buetti-Dinh, A., Galli, V., Bellenberg, S., Ilie, O., Herold, M., Christel, S., Boretska, M., Pivkin, I. V., Wilmes, P., Sand, W., Vera, M., \& Dopson, M. (2019). Deep neural networks outperform human expert’s capacity in characterizing bioleaching bacterial biofilm composition. Biotechnology Reports, 22, e00321. \url{https://doi.org/10.1016/j.btre.2019.e00321}\\

Christiaensen Luc, and Lionel Demery, eds. 2018.Agriculture in Africa: Telling Myths from Facts. Directions in Development. Washington, DC: World Bank. doi:10.1596/978-1-4648-1134-0. License: Creative Commons Attribution CC BY 3.0 IGO\\

Conway, G., Badiane, O., \& Glatzel, K. (2019). Food for All in Africa: Sustainable Intensification for African Farmers. Cornell University Press.\\

Gao, B.-. (1996). NDWI—A normalized difference water index for remote sensing of vegetation liquid water from space. Remote Sensing of Environment, 58(3), 257–266. \url{https://doi.org/10.1016/s0034-4257(96)00067-3}\\

Global Change Data Lab. (2021, February 24). Coronavirus Pandemic Data Explorer. Our World in Data. \url{https://ourworldindata.org/coronavirus-data-explorer?zoomToSelection=true}\\

Huang, J., \& Han, D. (2014). Meta-analysis of influential factors on crop yield estimation by remote sensing. International Journal of Remote Sensing, 35(6), 2267–2295. \url{https://doi.org/10.1080/01431161.2014.890761}\\

Idso, S. B., Jackson, R. D., \& Reginato, R. J. (1977). Remote-Sensing of Crop Yields. Science, 196(4285), 19–25. \url{https://doi.org/10.1126/science.196.4285.19}\\

IFC. (2020, September 4). COVID-19 Economic Impact: Sub-Saharan Africa. International Finance Corporation. \url{https://www.ifc.org/wps/wcm/connect/publications_ext_content/ifc_external_publication_site/publications_listing_page/covid-19-response-brief-ssa}\\

International Labor Organization. (2020, April). COVID-19 and the impact on agriculture and food security. ILO. \url{https://www.ilo.org/wcmsp5/groups/public/---ed_dialogue/---sector/documents/briefingnote/wcms_742023.pdf}\\

International Food Policy Research Institute (IFPRI); International Institute for Applied Systems Analysis (IIASA), 2016, “Global Spatially-Disaggregated Crop Production Statistics Data for 2005 Version 3.2”, \url{https://doi.org/10.7910/DVN/DHXBJX, Harvard Dataverse, V9}\\

International Food Policy Research Institute, 2019, “Global Spatially-Disaggregated Crop Production Statistics Data for 2010 Version 2.0”, \url{https://doi.org/10.7910/DVN/PRFF8V, Harvard Dataverse, V4}\\

International Food Policy Research Institute, 2020, “Spatially-Disaggregated Crop Production Statistics Data in Africa South of the Sahara for 2017”, \url{https://doi.org/10.7910/DVN/FSSKBW, Harvard Dataverse, V3}\\

Kaneko, A., Kennedy, T., Mei, L., Sintek, C., Burke, M., Ermon, S., \& Lobell, D. (2019). Deep Learning for Crop Yield Prediction in Africa.\\

Kpienbaareh, D., Sun, X., Wang, J., Luginaah, I., Bezner Kerr, R., Lupafya, E., \& Dakishoni, L. (2021). Crop Type and Land Cover Mapping in Northern Malawi Using the Integration of Sentinel-1, Sentinel-2, and PlanetScope Satellite Data. Remote Sensing, 13(4), 700. \url{https://doi.org/10.3390/rs13040700}\\

Lein, J. K. (2012). Environmental Sensing: Analytical Techniques for Earth Observation. New York, NY: Springer.\\

Leroux, L., Baron, C., Zoungrana, B., Traore, S. B., Lo Seen, D., \& Begue, A. (2016). Crop Monitoring Using Vegetation and Thermal Indices for Yield Estimates: Case Study of a Rainfed Cereal in Semi-Arid West Africa. IEEE Journal of Selected Topics in Applied Earth Observations and Remote Sensing, 9(1), 347–362. \url{https://doi.org/10.1109/jstars.2015.2501343}\\

Liu, J., Shang, J., Qian, B., Huffman, T., Zhang, Y., Dong, T., Jing, Q., \& Martin, T. (2019). Crop Yield Estimation Using Time-Series MODIS Data and the Effects of Cropland Masks in Ontario, Canada. Remote Sensing, 11(20), 2419. \url{https://doi.org/10.3390/rs11202419}\\

Li, Q., Qiu, C., Ma, L., Schmitt, M., \& Zhu, X. (2020). Mapping the Land Cover of Africa at 10 m Resolution from Multi-Source Remote Sensing Data with Google Earth Engine. Remote Sensing, 12(4), 602. \url{https://doi.org/10.3390/rs12040602}\\

Ly, R., \& Dia, K. (2020, August). https://akademiya2063.org/uploads/Covid-19-Bulletin-004.pdf (No. 004). AKADEMIYA2063. \url{https://akademiya2063.org/uploads/Covid-19-Bulletin-004.pdf}\\

Pekel, J.-F., Cottam, A., Gorelick, N., \& Belward, A. S. (2016). High-resolution mapping of global surface water and its long-term changes. Nature, 540(7633), 418–422. \url{https://doi.org/10.1038/nature20584}\\

Rasmussen, M. S. (1992). Assessment of millet yields and production in northern Burkina Faso using integrated NDVI from the AVHRR. International Journal of Remote Sensing, 13(18), 3431–3442. \url{https://doi.org/10.1080/01431169208904132}\\

Rasmussen, M. S. (1997). Operational yield forecast using AVHRR NDVI data: Reduction of environmental and inter-annual variability. International Journal of Remote Sensing, 18(5), 1059–1077. \url{https://doi.org/10.1080/014311697218575}\\

Rembold, F., Atzberger, C., Savin, I., \& Rojas, O. (2013). Using Low Resolution Satellite Imagery for Yield Prediction and Yield Anomaly Detection. Remote Sensing, 5(4), 1704–1733. \url{https://doi.org/10.3390/rs5041704}\\

Rezaei, E.E., Ghazaryan, G., González, J., Cornish, N., Dubovyk, O., \& Siebert, S. (2020). The use of remote sensing to derive maize sowing dates for large-scale crop yield simulations. Int J Biometeorol. \url{https://doi.org/10.1007/s00484-020-02050-4}\\

Running, S., Mu, Q., Zhao, M. (2021). MODIS/Terra Net Evapotranspiration 8-Day L4 Global 500m SIN Grid V061 [Data set]. NASA EOSDIS Land Processes DAAC. Accessed 2021-03-16 from \url{https://doi.org/10.5067/MODIS/MOD16A2.061}\\

Silver, D., Huang, A., Maddison, C. J., Guez, A., Sifre, L., van den Driessche, G., Schrittwieser, J., Antonoglou, I., Panneershelvam, V., Lanctot, M., Dieleman, S., Grewe, D., Nham, J., Kalchbrenner, N., Sutskever, I., Lillicrap, T., Leach, M., Kavukcuoglu, K., Graepel, T., \& Hassabis, D. (2016). Mastering the game of Go with deep neural networks and tree search. Nature, 529(7587), 484–489. \url{https://doi.org/10.1038/nature16961}\\

Smith, R. C. G., Barrs, H. D., Steiner, J. L., \& Stapper, M. (1985). Relationship between wheat yield and foliage temperature: theory and its application to infrared measurements. Agricultural and Forest Meteorology, 36(2), 129–143. \url{https://doi.org/10.1016/0168-1923(85)90005-x}\\

Stockholm International Water Institute (SIWI). (2018). Unlocking the potential of enhanced rainfed agriculture (No. 39). \url{https://www.siwi.org/wp-content/uploads/2018/12/Unlocking-the-potential-of-rainfed-agriculture-2018-FINAL.pdf}\\

Volodymyr Mnih, Koray Kavukcuoglu, David Silver, Alex Graves, Ioannis Antonoglou, Daan Wierstra, and Martin Riedmiller. Playing atari with deep reinforcement learning. arXiv preprint arXiv:1312.5602, 2013\\

Wan, Z., Hook, S., Hulley, G. (2015). MOD11A2 MODIS/Terra Land Surface Temperature/Emissivity 8-Day L3 Global 1km SIN Grid V006 [Data set]. NASA EOSDIS Land Processes DAAC. Accessed 2021-03-16 from \url{https://doi.org/10.5067/MODIS/MOD11A2.006}\\

Zeufack, Albert G.; Calderon, Cesar; Kambou, Gerard; Djiofack, Calvin Z.; Kubota, Megumi; Korman, Vijdan; Cantu Canales, Catalina. 2020. "Africa’s Pulse, No. 21" (April), World Bank, Washington, DC. Doi: 10.1596/978-1-4648-1568-3. License: Creative Commons Attribution CC BY 3.0 IGO.

\newpage

\begin{center}
\begin{Huge}
\vspace*{290pt}
Appendices
\end{Huge}
\end{center}

\newpage

\begin{landscape}
\begin{table}[!htbp]
\caption {Most significant commodities by country in western Africa region based on our selection criteria.} \label{tab:table5}
\begin{tabular}{llllll}
Country & Commodity 1 & Commodity 2 & Commodity 3 & Commodity 4   & Commodity 5 \\ 
\hline
Benin           & Cassava   and products    & Yams                         & Maize   and products      & Beverages,   Fermented & Palm   kernels               \\
Burkina   Faso  & Beverages,   Fermented    & Sorghum   and products       & Maize   and products      & Millet   and products  & Pulses,   Other and products \\
Cabo   Verde    & Sugar   cane              & Pelagic   Fish               & Tomatoes   and products   & Vegetables,   Other    & Milk   - Excluding Butter    \\
Côte   d’Ivoire & Yams                      & Cassava   and products       & Rice   and products       & Palm   kernels         & Sugar   cane                 \\
The   Gambia    & Groundnuts   (Shelled Eq) & Millet   and products        & Milk   - Excluding Butter & Rice   and products    & Beverages,   Fermented       \\
Ghana           & Cassava   and products    & Yams                         & Plantains                 & Palm   kernels         & Maize   and products         \\
Guinea          & Rice   and products       & Cassava   and products       & Palm   kernels            & Maize   and products   & Groundnuts   (Shelled Eq)    \\
Guinea-Bissau   & Rice   and products       & Nuts   and products          & Roots,   Other            & Palm   kernels         & Plantains                    \\
Liberia         & Cassava   and products    & Rice   and products          & Sugar   cane              & Palm   kernels         & Bananas                      \\
Mali            & Maize   and products      & Rice   and products          & Millet   and products     & Vegetables,   Other    & Milk   - Excluding Butter    \\
Mauritania      & Pelagic   Fish            & Milk   - Excluding Butter    & Rice   and products       & Demersal   Fish        & Sorghum   and products       \\
Niger           & Millet   and products     & Pulses,   Other and products & Sorghum   and products    & Vegetables,   Other    & Milk   - Excluding Butter    \\
Nigeria         & Cassava   and products    & Yams                         & Vegetables,   Other       & Maize   and products   & Palm   kernels               \\
Senegal         & Sugar   cane              & Groundnuts   (Shelled Eq)    & Rice   and products       & Millet   and products  & Vegetables,   Other          \\
Sierra   Leone  & Cassava   and products    & Rice   and products          & Vegetables,   Other       & Palm   kernels         & Milk   - Excluding Butter    \\
Togo            & Cassava   and products    & Maize   and products         & Yams                      & Sorghum   and products & Beans                       
\end{tabular}
\end{table}

\begin{table}[!htbp]
\caption {Most significant commodities by country in northern Africa region based on our selection criteria.} \label{tab:table6}
\begin{tabular}{llllll}
Country & Commodity   1        & Commodity   2             & Commodity   3             & Commodity   4             & Commodity   5                  \\
\hline
Algeria & Vegetables,   Other  & Potatoes   and products   & Milk   - Excluding Butter & Wheat   and products      & Onions                         \\
Egypt   & Sugar   cane         & Sugar   beet              & Wheat   and products      & Vegetables,   Other       & Maize   and products           \\
Morocco & Wheat   and products & Sugar   beet              & Vegetables,   Other       & Milk   - Excluding Butter & Barley   and products          \\
Sudan   & Sugar   cane         & Sorghum   and products    & Milk   - Excluding Butter & Groundnuts   (Shelled Eq) & Onions                         \\
Tunisia & Vegetables,   Other  & Milk   - Excluding Butter & Tomatoes   and products   & Wheat   and products      & Olives   (including preserved)
\end{tabular}
\end{table}

\begin{table}[!htbp]
\caption {Most significant commodities by country in southern Africa region based on our selection criteria.} \label{tab:table7}
\begin{tabular}{llllll}
Country        & Commodity   1             & Commodity   2            & Commodity   3             & Commodity   4        & Commodity   5             \\
\hline
Botswana       & Milk   - Excluding Butter & Beer                     & Roots,   Other            & Vegetables,   Other  & Bovine   Meat             \\
Eswatini       & Sugar   cane              & Sugar   (Raw Equivalent) & Alcohol,   Non-Food       & Maize   and products & Roots,   Other            \\
Lesotho        & Milk   - Excluding Butter & Potatoes   and products  & Maize   and products      & Beer                 & Vegetables,   Other       \\
Namibia        & Roots,   Other            & Pelagic   Fish           & Beer                      & Demersal   Fish      & Milk -   Excluding Butter \\
South   Africa & Sugar   cane              & Maize   and products     & Milk   - Excluding Butter & Beer                 & Potatoes   and products  
\end{tabular}
\end{table}

\newpage

\begin{table}[!htbp]
\caption {Most significant commodities by country in central Africa region based on our selection criteria.} \label{tab:table8}
\begin{tabular}{llllll}
Country                      & Commodity 1          & Commodity 2             & Commodity 3             & Commodity 4             & Commodity 5    \\
\hline
Angola                       & Cassava and products & Bananas                 & Maize and products      & Sweet potatoes          & Beer           \\
Cameroon                     & Cassava and products & Plantains               & Maize and products      & Palm kernels            & Roots, Other   \\
The Central African Republic & Cassava and products & Yams                    & Groundnuts (Shelled Eq) & Roots, Other            & Sugar cane     \\
Chad                         & Sorghum and products & Groundnuts (Shelled Eq) & Millet and products     & Milk - Excluding Butter & Cereals, Other \\
Congo                        & Cassava and products & Sugar cane              & Beer                    & Vegetables, Other       & Palm kernels   \\
Gabon                        & Plantains            & Cassava and products    & Sugar cane              & Yams                    & Beer           \\
Sao Tome and Principe        & Plantains            & Coconuts - Incl Copra   & Palm kernels            & Roots, Other            & Pelagic Fish  
\end{tabular}
\end{table}

\begin{table}[!htbp]
\caption {Most significant commodities by country in eastern Africa region based on our selection criteria.} \label{tab:table9}
\begin{tabular}{llllll}
Country                     & Commodity 1          & Commodity 2             & Commodity 3            & Commodity 4             & Commodity 5             \\
\hline
Djibouti                    & Vegetables, Other    & Milk - Excluding Butter & Bovine Meat            & Mutton \& Goat Meat     & Fruits, Other           \\
Ethiopia                    & Maize and products   & Roots, Other            & Cereals, Other         & Sorghum and products    & Wheat and products      \\
Kenya                       & Sugar cane           & Milk - Excluding Butter & Maize and products     & Vegetables, Other       & Potatoes and products   \\
Madagascar                  & Rice and products    & Sugar cane              & Cassava and products   & Sweet potatoes          & Fruits, Other           \\
Malawi                      & Cassava and products & Sweet potatoes          & Maize and products     & Sugar cane              & Fruits, Other           \\
Mauritius                   & Sugar cane           & Sugar (Raw Equivalent)  & Vegetables, Other      & Poultry Meat            & Beer                    \\
Mozambique                  & Cassava and products & Sugar cane              & Maize and products     & Milk - Excluding Butter & Bananas                 \\
Comoros                     & Pelagic Fish         & Marine Fish, Other      & Demersal Fish          & Crustaceans             &                         \\
Rwanda                      & Bananas              & Sweet potatoes          & Cassava and products   & Potatoes and products   & Plantains               \\
Seychelles                  & Pelagic Fish         & Demersal Fish           & Marine Fish, Other     & Fish, Body Oil          & Aquatic Animals, Others \\
Uganda                      & Sugar cane           & Plantains               & Cassava and products   & Maize and products      & Beverages, Fermented    \\
United Republic of Tanzania & Maize and products   & Cassava and products    & Sweet potatoes         & Bananas                 & Sugar cane              \\
Zambia                      & Sugar cane           & Maize and products      & Cassava and products   & Sugar (Raw Equivalent)  & Milk - Excluding Butter \\
Zimbabwe                    & Sugar cane           & Maize and products      & Sugar (Raw Equivalent) & Milk - Excluding Butter & Cassava and products   
\end{tabular}
\end{table}

\end{landscape}

\begin{figure}[htbp!]
  \caption{The 2020 predicted rice production in west African countries. \textbf{Note}: Data, methodology, and maps’ sources: Authors.}
  \centering
    \includegraphics[width=0.75\textwidth]{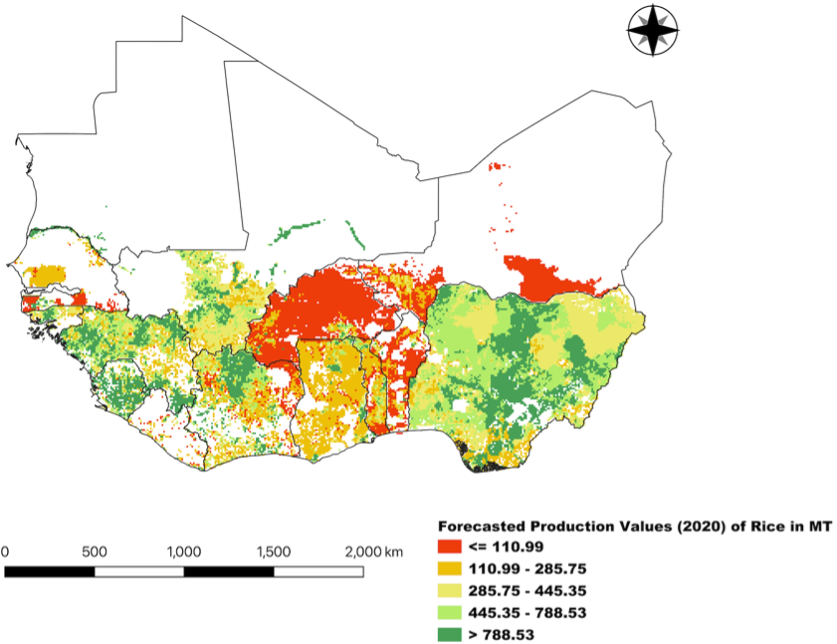}
\end{figure}
    
\begin{figure}[htbp!]
  \caption{The 2020 predicted maize production in west African countries. \textbf{Note}: Data, methodology, and maps’ sources: Authors.}
  \centering
    \includegraphics[width=0.75\textwidth]{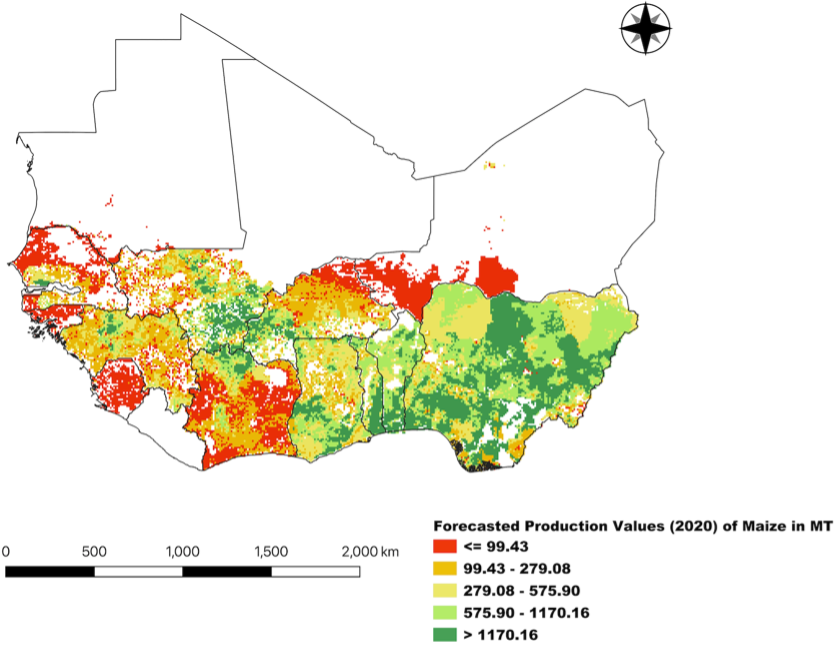}
\end{figure}

\begin{figure}[htbp!]
  \caption{The 2020 predicted maize production in west African countries. \textbf{Note}: Data, methodology, and maps’ sources: Authors.}
  \centering
    \includegraphics[width=0.75\textwidth]{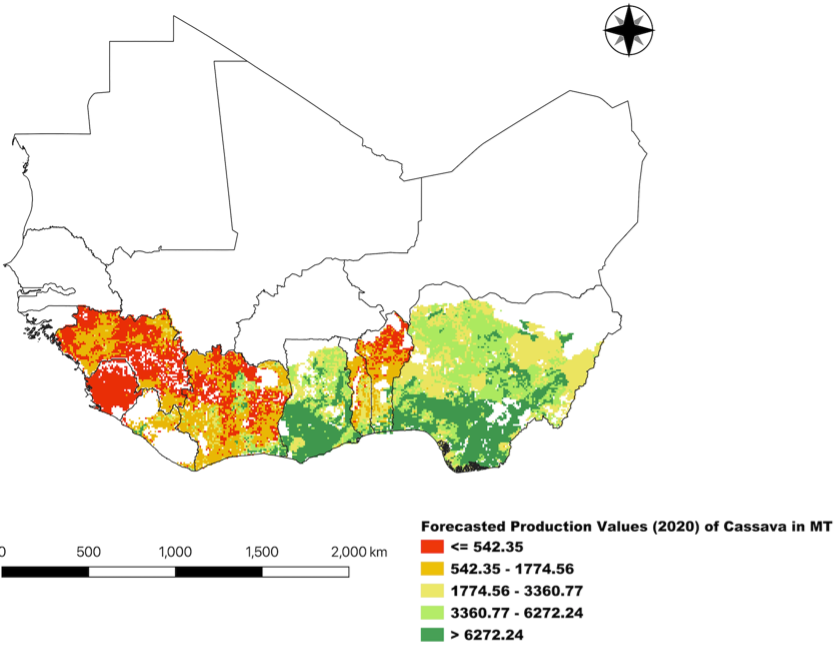}
\end{figure}

\begin{figure}[htbp!]
  \caption{The 2020 predicted maize production in southern African countries. \textbf{Note}: Data, methodology, and maps’ sources: Authors.}
  \centering
    \includegraphics[width=0.75\textwidth]{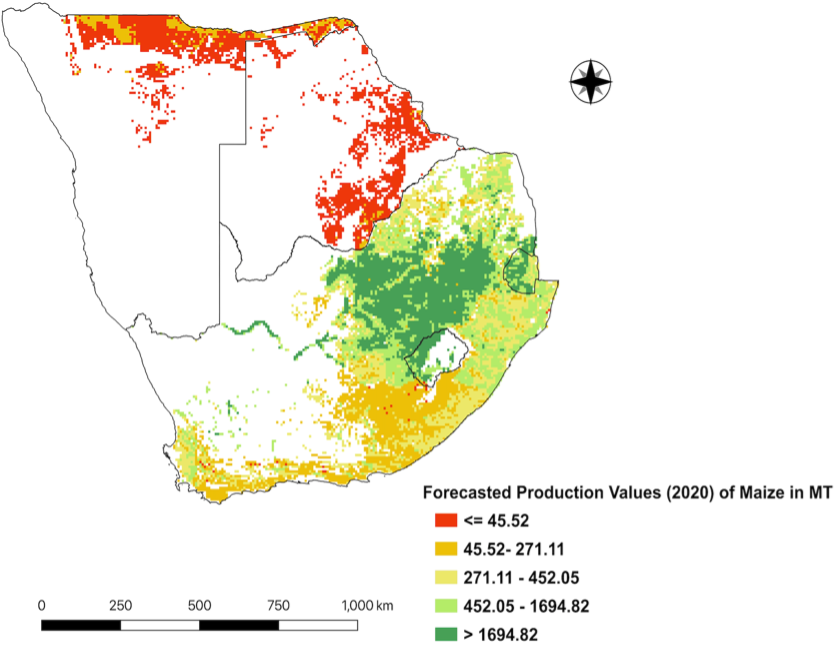}
\end{figure}

\begin{figure}[htbp!]
  \caption{The 2020 predicted maize production in northern African countries. \textbf{Note}: Data, methodology, and maps’ sources: Authors.}
  \centering
    \includegraphics[width=0.75\textwidth]{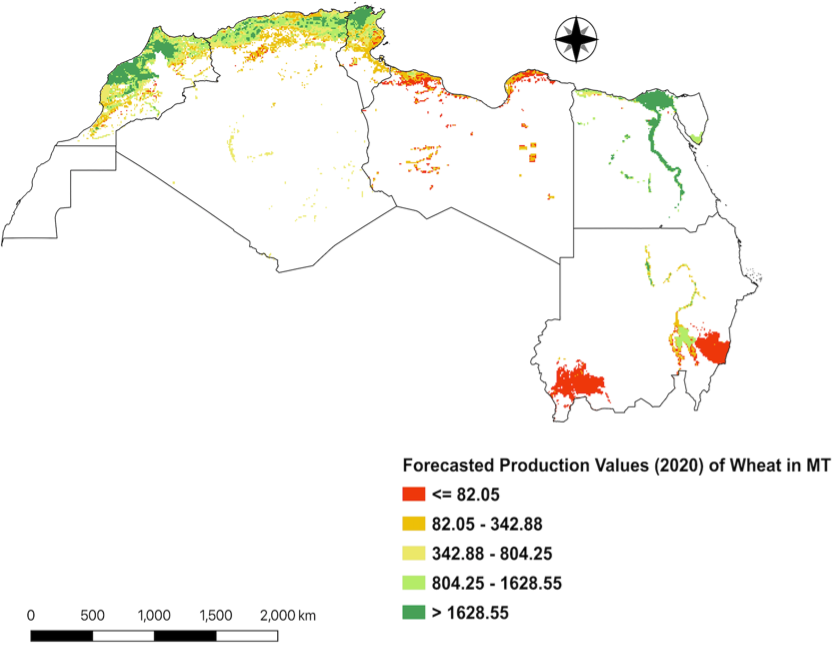}
\end{figure}

\begin{figure}[htbp!]
  \caption{The 2020 predicted maize production in eastern African countries. \textbf{Note}: Data, methodology, and maps’ sources: Authors.}
  \centering
    \includegraphics[width=0.75\textwidth]{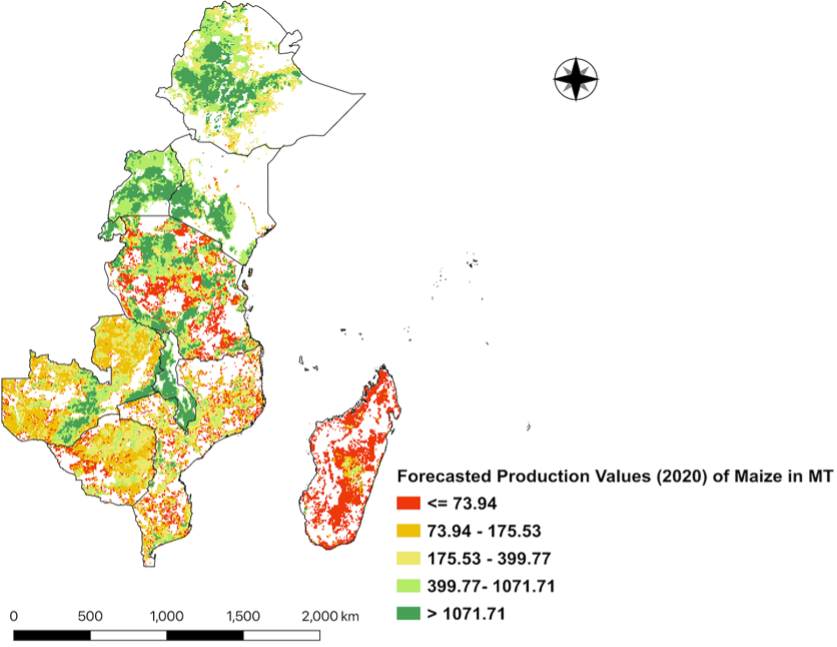}
\end{figure}

\begin{figure}[htbp!]
  \caption{The 2020 predicted maize production in central African countries. \textbf{Note}: Data, methodology, and maps’ sources: Authors.}
  \centering
    \includegraphics[width=0.75\textwidth]{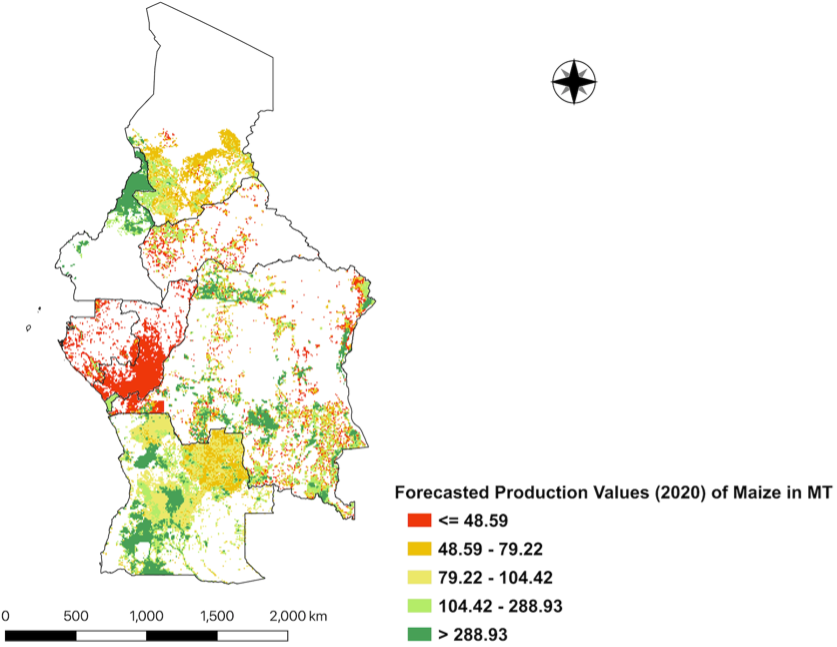}
\end{figure}

\end{document}